\documentclass[10pt,twocolumn,letterpaper]{article}

\usepackage{cvpr}              

\usepackage{graphicx}
\usepackage{amsmath}
\usepackage{amssymb}
\usepackage{booktabs}

\usepackage[pagebackref,breaklinks,colorlinks]{hyperref}

\usepackage[capitalize]{cleveref}
\crefname{section}{Sec.}{Secs.}
\Crefname{section}{Section}{Sections}
\Crefname{table}{Table}{Tables}
\crefname{table}{Tab.}{Tabs.}

\def\cvprPaperID{*****} 
\def\confName{CVPR}
\def\confYear{2022}

\begin{document}

\title{E-Scooter Rider Detection and Classification in Dense Urban Environments}

\author{Shane Gilroy\textsuperscript{1,2}, Darragh Mullins\textsuperscript{1}, Edward Jones\textsuperscript{1}, Ashkan Parsi\textsuperscript{1} and Martin Glavin\textsuperscript{1}\\
\textsuperscript{1}National University of Ireland, Galway, Ireland\\
\textsuperscript{2}Atlantic Technological University, Ireland \\
}
\maketitle

\begin{abstract}
   
  Accurate detection and classification of vulnerable road users is a safety critical requirement for the deployment of autonomous vehicles in heterogeneous traffic. Although similar in physical appearance to pedestrians, e-scooter riders follow distinctly different characteristics of movement and can reach speeds of up to 45kmph. The challenge of detecting e-scooter riders is exacerbated in urban environments where the frequency of partial occlusion is increased as riders navigate between vehicles, traffic infrastructure and other road users. This can lead to the non-detection or mis-classification of e-scooter riders as pedestrians, providing inaccurate information for accident mitigation and path planning in autonomous vehicle applications. This research introduces a novel benchmark for partially occluded e-scooter rider detection to facilitate the objective characterization of detection models. A novel, occlusion-aware method of e-scooter rider detection is presented that achieves a 15.93\% improvement in detection performance over the current state of the art.
  
   
\end{abstract}


\section{Introduction}
\label{sec:intro}

Accurate detection and classification of vulnerable road users (pedestrians, cyclists, and micro-mobility users) is a safety critical requirement for the roll out of autonomous vehicles in heterogeneous traffic.
The SAE J3016 standard \cite{sae2018taxonomy} defines levels of driving automation ranging from Level 0, where the vehicle contains zero automation and the human driver is in complete control, to level 5 where the vehicle is solely responsible for all perception and driving tasks in all scenarios. The progression from automation levels 3-5 requires a significant increase in assumption of responsibility by the vehicle, placing progressively increasing demands on the performance of detection and classification systems to inform efficient path planning, accident mitigation, and to ensure the safety of vulnerable road users.
Despite recent improvements in detection systems, many challenges still exist before we reach the object detection capabilities required for safe autonomous driving in urban environments.
One of the most complex and persistent challenges is that of partial occlusion, where a target object is only partially available to the sensor due to obstruction by another foreground object.

Micro-mobility solutions such as e-scooters have seen a rapid rise in popularity in recent years as many cities seek modern solutions to ease traffic, emissions and parking difficulties in built up areas. The intuitive operation of e-scooters, and the growing number of service providers offering short term rentals, have prompted market predictions that shared e-scooter usage may ultimately capture 8-15\% of all trips shorter than 5 miles, worldwide \cite{heineke2019micromobility}.
This proliferation of e-scooters usage adds an additional level of complexity to the detection and classification of vulnerable road users. Although very similar in physical appearance, e-scooter riders and pedestrians behave very differently in the automotive environment. E-scooter riders can reach speeds of up to 45 kilometers per hour \cite{ewert2021small}\cite{hardt2019usage}\cite{european2013regulation} and follow distinctly different movement characteristics than pedestrians. The challenge of accurately detecting and classifying e-scooter riders is exacerbated in urban environments where the frequency and severity of partial occlusion is increased as vulnerable road users (VRUs) navigate between vehicles, buildings, traffic infrastructure, other road users. This can lead to the non-detection or mis-classification of e-scooter riders as pedestrians or other road users, providing inaccurate information for accident mitigation and path planning.
In addition, recent research indicates that e-scooter usage is currently one of the most dangerous forms of transportation with 115 injuries per million trips \cite{ioannides2022scooter}, substantially higher than motorcycles (104 injuries per million trips), bicycles (15 injuries per million trips) and walking (2 injuries per million trips) \cite{beck2007motor}. 

Leading pedestrian and cyclist detection systems claim a detection performance of approximately 65\%-75\% of partially and heavily occluded instances respectively using current benchmarks \cite{cao2021handcrafted}\cite{gilroy2021pedestrian}\cite{gilroy2019overcoming}\cite{ning2021survey}\cite{xiao2021deep}. However, limited work has been carried out on the safety critical challenge of e-scooter rider detection to date and to the best of the authors knowledge, no known research has been carried out on the detection and classification of e-scooter riders under partial occlusion.

This research uses a novel, objective benchmark for partially occluded e-scooter riders to facilitate the characterization of vulnerable road user detection and classification models. A novel, occlusion-aware method of e-scooter rider detection is presented and objective performance characterization is carried out for a range of popular classifiers for the complete spectrum of occlusion levels from 0-99\%.
The contributions of this research are:
1. A novel, objective, test benchmark for partially occluded e-scooter rider detection and classification is presented.
2. A novel, occlusion-aware method of e-scooter rider detection is described which provides a 15.93\% improvement on the current state of the art e-scooter rider detection network as described in \cite{apurv2021detection}.
3. Objective characterization of e-scooter rider classification is carried out for a number of popular, publicly available classifiers.




\section{Related Work}
\label{sec:Section2}
Limited research has been carried out on e-scooter rider detection to date.
Apurv \textit{et al} \cite{apurv2021detection} present a baseline algorithm for e-scooter rider detection. Candidate selection is carried out using YoloV3 with pre-trained weights \cite{redmon2018yolov3} on the COCO dataset \cite{lin2014microsoft}. The bounding box dimensions for each person instance are then enlarged on three sides (left, bottom and right) using the formula outlined in Equation \ref{eq:1}, to incorporate the surrounding area where an e-scooter is normally located in instances where the detected person is an e-scooter rider.

\begin{equation}\label{eq:1}
(x,y,w,h) = ((x-w),y,3w,(h+h/4))
\end{equation}

The extended bounding box regions are then fed into a MobileNetV2 classifier \cite{sandler2018mobilenetv2}, trained on the "IUPUI CSRC E-Scooter Rider Detection Benchmark Dataset" \cite{apurv2021detection}. The IUPUI E-Scooter Rider Dataset contains 21,454 images for binary classification including 10,749 images containing e-scooter riders and 10,705 images which do not contain an e-scooter rider. The authors claim a validation accuracy of more than 0.9, however very few instances of occluded e-scooter riders are included in the validation data and no reference is made as to the ability of the network to generalize to new data.

Nguyen \textit{et al} \cite{nguyen2021electric} also utilise YoloV3 to implement an e-scooter rider detection system, however this approach focuses on detecting an e-scooter and its rider as two separate classes. The methodology separates the image into an even grid and relates parallel bounding boxes of the target classes in order to identify e-scooter riders. The network is trained using 140 training images and 60 validation images obtained through web trawling on Baidu and Google Images using the keyword “rider and scooter”. Transfer learning is then used to fine tune the YoloV3 model to the target classes. The authors expand this research by exploiting the detection of two separate classes to identify cases where the detected person is horizontal to the e-scooter, indicating a potential fall. The authors also claim a validation accuracy of over 0.9, however only 60 validation images are used, no instances of occluded e-scooter riders are included and no reference is made to more thorough evaluation indicating the networks’ ability to generalize to new data. 

Researchers at the Digital Transformation Hub at California Polytechnic State University collaborated with the City of Santa Monica in 2018 to implement a machine learning based e-scooter detection and counting system, in order to help monitor and enforce the prevention of e-scooter riding on sidewalks\cite{git_cal_poly2021}. E-scooter rider detection was achieved through transfer learning of a pre-trained RetinaNet object detection algorithm using an in-house custom dataset. A parallel Resnet50 semantic segmentation branch was also used to differentiate between the sidewalk and the road surface. Overlapping e-scooter rider and sidewalk detections indicate an infringement and the instance is counted and tracked for enforcement purposes \cite{cal_poly2018}.

Many popular pedestrian detection benchmarks provide occlusion level annotation to determine the relative detection performance for partially occluded pedestrians\cite{braun2019eurocity}\cite{chi2020pedhunter}\cite{choi2018kaist}\cite{dollar2011pedestrian}\cite{geiger2012we}\cite{hwang2015multispectral}\cite{li2016new}\cite{li2016unified}\cite{pang2020tju}\cite{shao2018crowdhuman}\cite{zhang2016far}\cite{zhang2017citypersons}. Although less represented, there is also a significant number of cyclist detection benchmarks with occlusion specific annotation \cite{braun2019eurocity}\cite{choi2018kaist}\cite{geiger2012we}\cite{li2016unified}, however, no known e-scooter detection benchmark with occlusion labels exists to date. 


Gilroy \textit{et al} \cite{gilroy2022impact} present an objective benchmark for partially occluded pedestrian detection, containing 820 pedestrian instances under progressive levels of occlusion from 0-99\%. Images are annotated using the objective method of occlusion level annotation described in \cite{gilroy2022objective}. Keypoint detection is used to identify semantic body parts and findings are cross-referenced with a visibility score and the pedestrian mask in order to confirm the presence or occlusion of each semantic part. A novel method of 2D body surface area estimation based on the "Wallace rule of Nines" \cite{gilroy2021pedestrian}\cite{wallace1951exposure} is then used to calculate the total occlusion level of each pedestrian instance. Inspired by the work of Gilroy \textit{et al} \cite{gilroy2022impact}, this research uses a novel objective benchmark for e-scooter rider detection and proposes a novel, occlusion-aware e-scooter rider detection network to improve upon the current state of the art.




\section{Methodology}

\begin{figure}[]
\begin{center}
   \includegraphics[width=\linewidth]{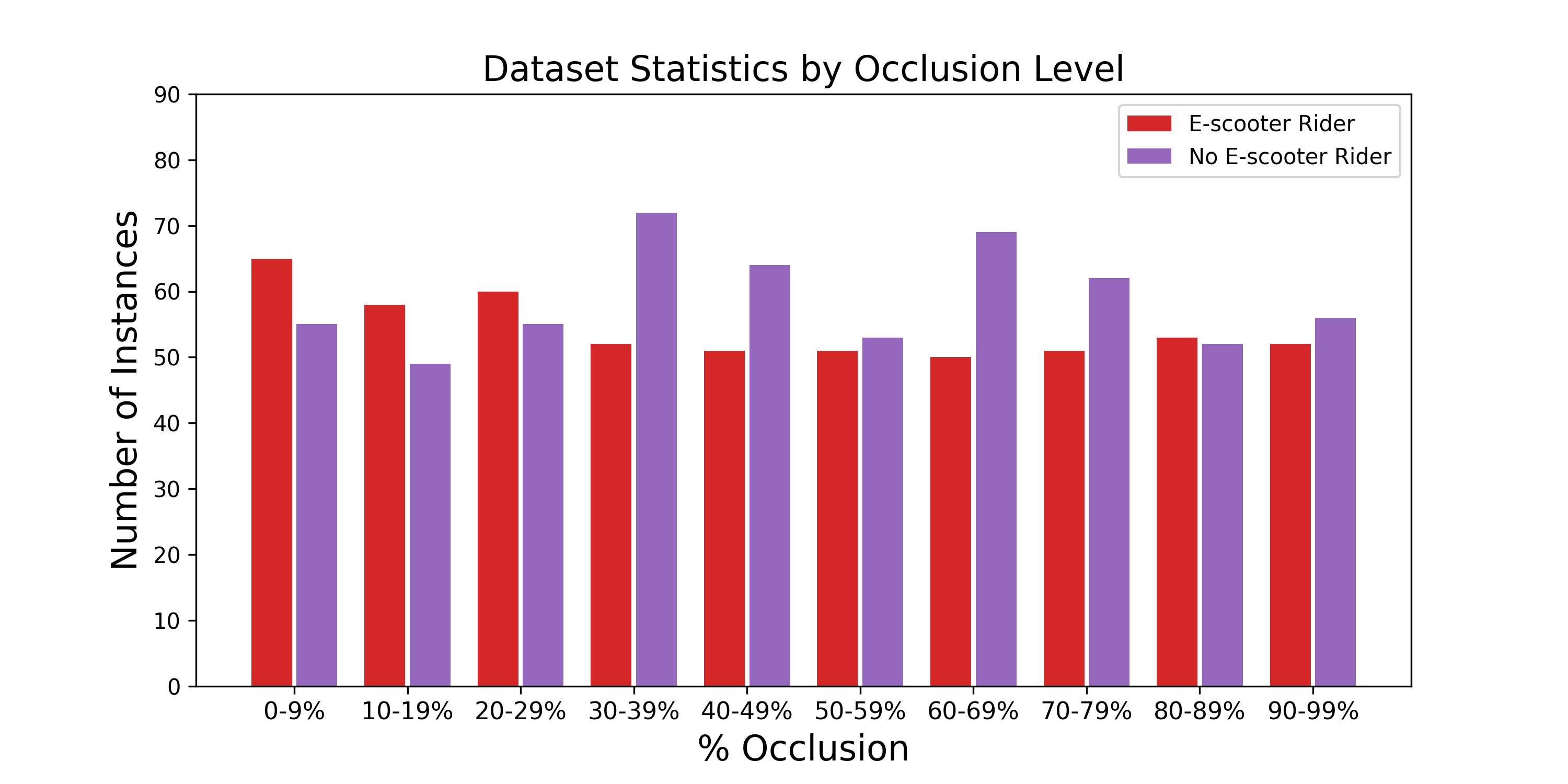}
\end{center}
  \caption{Test Dataset Statistics. The number of target instances per occlusion level. The custom dataset contains 1,130 images under progressive levels of occlusion from 0-99\%.}
\label{fig:ds_stats}
\end{figure}

\begin{figure}[]
\begin{center}
   \includegraphics[width=\linewidth]{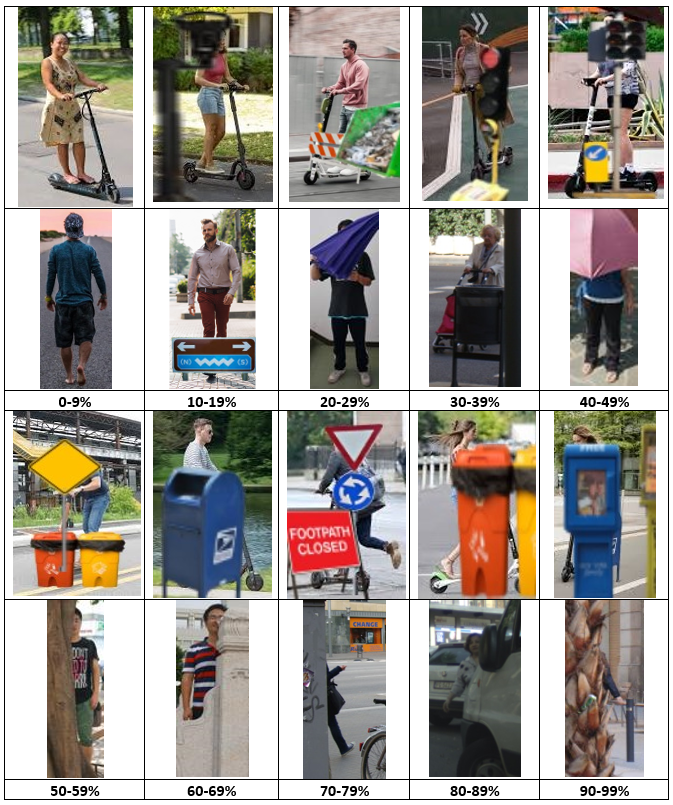}
\end{center}
  \caption{Test Dataset Sample. An example of dataset images for each level of occlusion. The custom dataset consists of 1,130 images, including 543 e-scooter rider images and 587 non e-scooter rider images. Images include a diverse mix of natural and superimposed occlusions, and contain a wide range of poses, orientations and occluding objects. All images are compiled from publicly available sources.}
\label{fig:ds}
\end{figure}

A novel e-scooter rider test dataset, containing 1,130 images including 543 e-scooter rider instances and 587 other vulnerable road user instances, has been created in order to characterize e-scooter rider detection and classification across a range of occlusion levels from 0 to 99\% occluded. 
A diverse mix of images are used ensure that a wide variety of e-scooter riders, orientations, backgrounds, and occluding objects are represented. The dataset is compiled from publicly available, web crawled sources. Occluding objects are superimposed on to e-scooter riders with progressive levels of occlusion. This dataset is then complemented by 587 instances of pedestrians and cyclists across an identical range of occlusion levels. Non e-scooter rider images are collated from multiple publicly available sources including \cite{gilroy2021pedestrian}\cite{braun2019eurocity}\cite{zhang2017citypersons}\cite{gilroy2022impact}\cite{zheng2015partial}\cite{zhuo2018occluded}.
All images are annotated using the objective occlusion level classification method described in \cite{gilroy2022objective}. Complex cases at very high occlusion rates were manually verified using the method of 2D body surface area estimation presented in \cite{gilroy2021pedestrian}.
Dataset statistics by occlusion level and a sample of the test dataset can be seen in Figure \ref{fig:ds_stats} and Figure \ref{fig:ds} respectively.


\subsection{E-Scooter Rider Classification}
Classifier performance is evaluated using the total test dataset for the current state of the art, as outlined in \cite{apurv2021detection}, and for five popular, publicly available classifiers in order to compare performance across the complete range of occlusion levels. Each classifier, AlexNet \cite{krizhevsky2014one}, SqueezNet1.0 \cite{iandola2016squeezenet}, VGG16 with Batch Normalisation (VGG16\_bn) \cite{simonyan2014very}, ResNet34 and ResNet101 \cite{he2016deep} is trained on the IUPUI E-Scooter Rider Dataset \cite{apurv2021detection} using Pytorch \cite{paszke2019pytorch} and Fast AI \cite{howard2020fastai}. The detection and classification pipeline proposed in \cite{apurv2021detection} is used to maintain consistency and provide baseline results for comparison.

Analysis is carried out using Voxel51\cite{moore2020fiftyone} and COCO style evaluation metrics. Accuracy is calculated using the formula highlighted in Equation \ref{eq:2}, where $TP$ = Number of true positives, $TN$ = Number of true negatives, $FP$ = Number of false positives and $FN$ = Number of false negatives.

\begin{equation}\label{eq:2}
Accuracy = \frac{TP + TN}{TP + TN + FP + FN}
\end{equation}

\begin{figure}[]
\begin{center}
   \includegraphics[width=\linewidth]{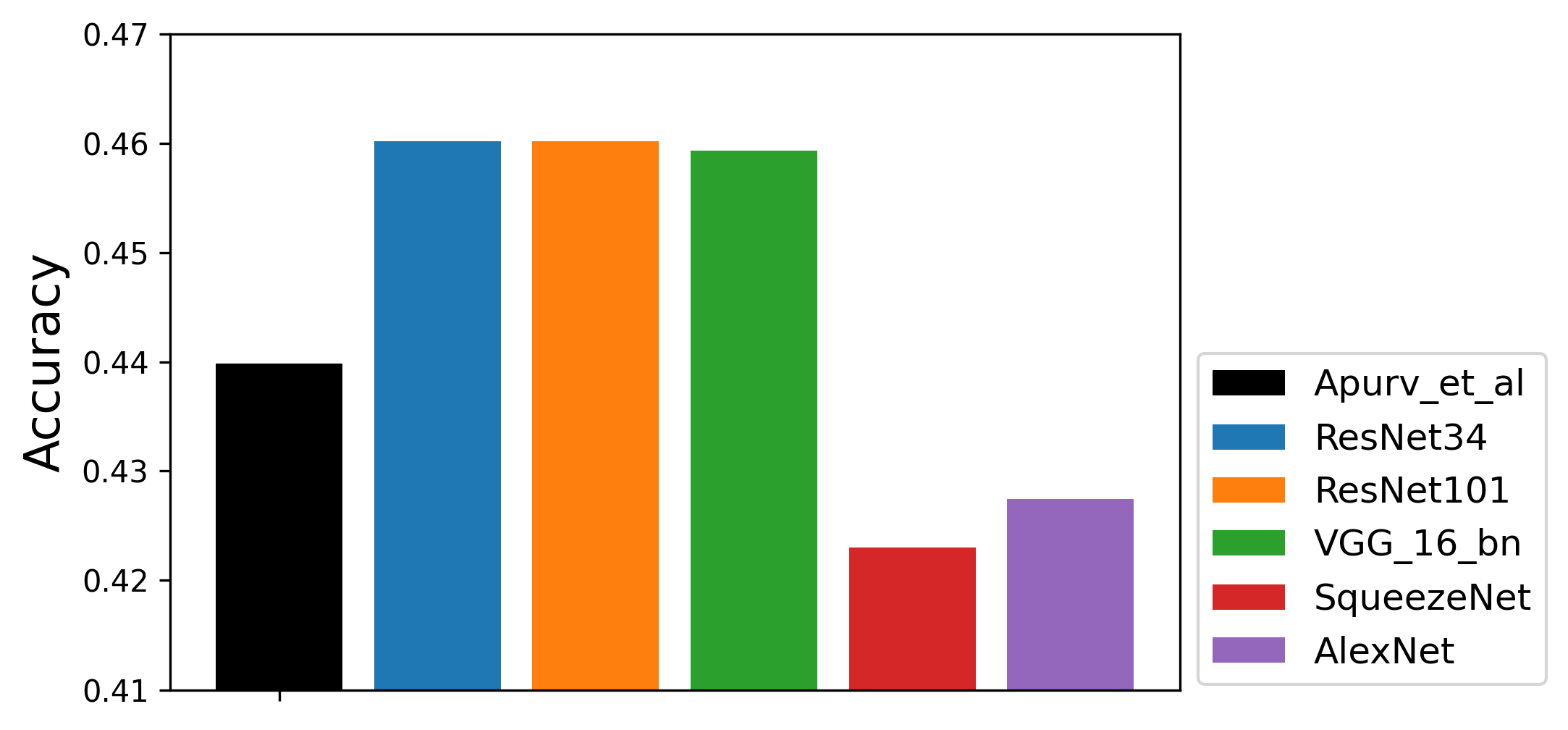}
\end{center}
  \caption{Classifier Performance. Classifier performance for the total test dataset using the methodology outlined by the current state of the art \cite{apurv2021detection}. ResNet101 and ResNet34 are the highest performing classifiers, each with an test accuracy of 0.460. The baseline method proposed by Apurv \textit{et al} \cite{apurv2021detection} has a test accuracy of 0.439.}
\label{fig:yoloP}
\end{figure}

A comparison of classifier performance using the methodology outlined in \cite{apurv2021detection} can be seen in Figure \ref{fig:yoloP}.


\subsection{Occlusion-Aware E-Scooter Rider Detection}
\label{sec:occ}

\begin{figure}[]
\begin{center}
   \includegraphics[width=\linewidth]{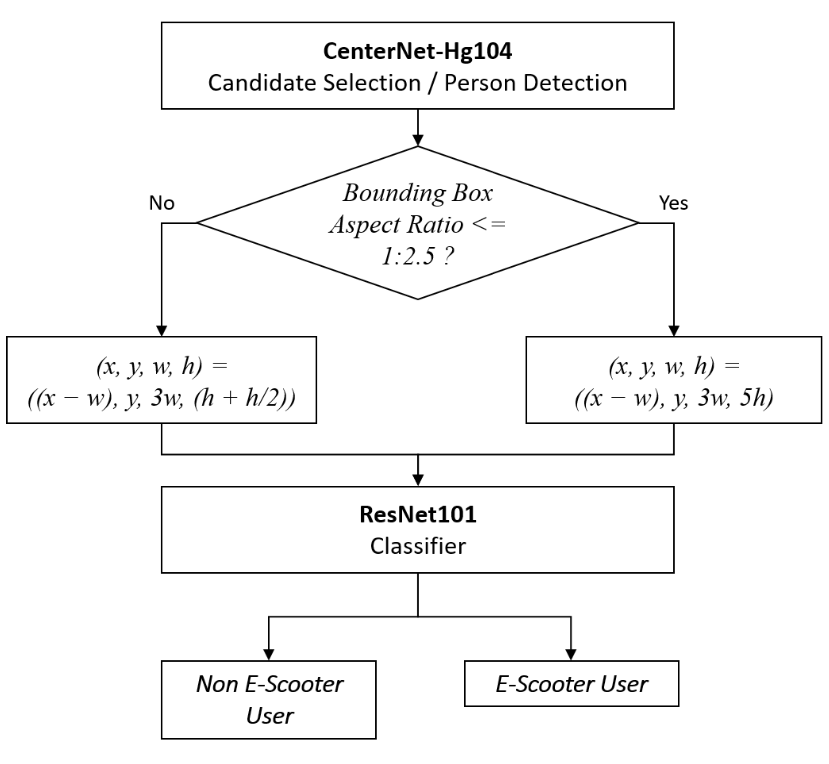}
\end{center}
  \caption{Occlusion-Aware E-Scooter Detection Flowchart.}
\label{fig:flow}
\end{figure}

\begin{figure}[]
\begin{center}
   \includegraphics[width=\linewidth]{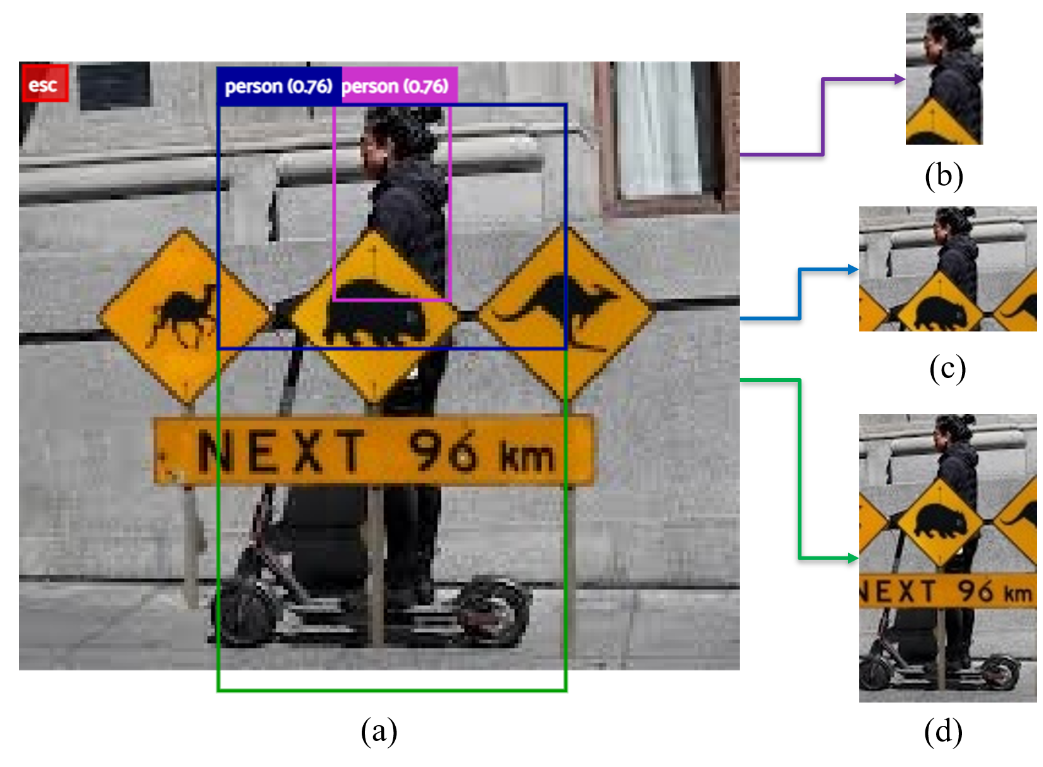}
\end{center}
  \caption{Candidate Selection Output Comparison. Example of the efficacy of the proposed candidate selection output for partially occluded e-scooter riders. The input image is displayed in (a). The cropped bounding box area from the initial detection algorithm is shown in (b). The cropped bounding box from the current state of the art as presented in \cite{apurv2021detection}, is shown in (c). The cropped bounding box area for the proposed novel, occlusion-aware e-scooter rider detection method is displayed in (d). The proposed method more comprehensively incorporates the e-scooter for partially occluded instances than the prior state of the art.}
\label{fig:pipline_crop}
\end{figure}

A novel, occlusion-aware method of e-scooter rider detection is proposed to increase the performance of e-scooter detection in heterogeneous traffic. Potential e-scooter rider instances are detected using a CenterNet-Hourglass104 \cite{zhou2019objects} based, COCO trained person detector. The aspect ratio of each bounding box is then analysed to determine if the detected instance is likely to be occluded. The detected bounding boxes of all potential candidates are then expanded on 3 sides as outlined in Figure \ref{fig:flow}. The extent to which the bounding boxes are expanded is based on the aspect ratio of the initial detection. If the bounding box height is less than 2.5 times the bounding box width, the person is more likely to be occluded and the height of the bounding box is increased by a higher magnitude to incorporate the pixel area where an e-scooter would be located in normal operation. An example of the efficacy of this method for partially occluded e-scooter users, compared to the current state of the art, can be seen in Figure \ref{fig:pipline_crop}.
The modified bounding boxes are then processed by a custom trained ResNet101 classifier. The classifier is trained using the e-scooter rider dataset presented in \cite{apurv2021detection}. The dataset contains 21,454 training images for binary classification, consisting of 10,749 "e-scooter rider" images and 10,705 "non e-scooter rider" images. A flowchart of the proposed occlusion-aware detection pipeline can be seen in Figure \ref{fig:flow}.


\subsection{Performance Characterization}
\begin{figure}[]
\begin{center}
   \includegraphics[width=\linewidth]{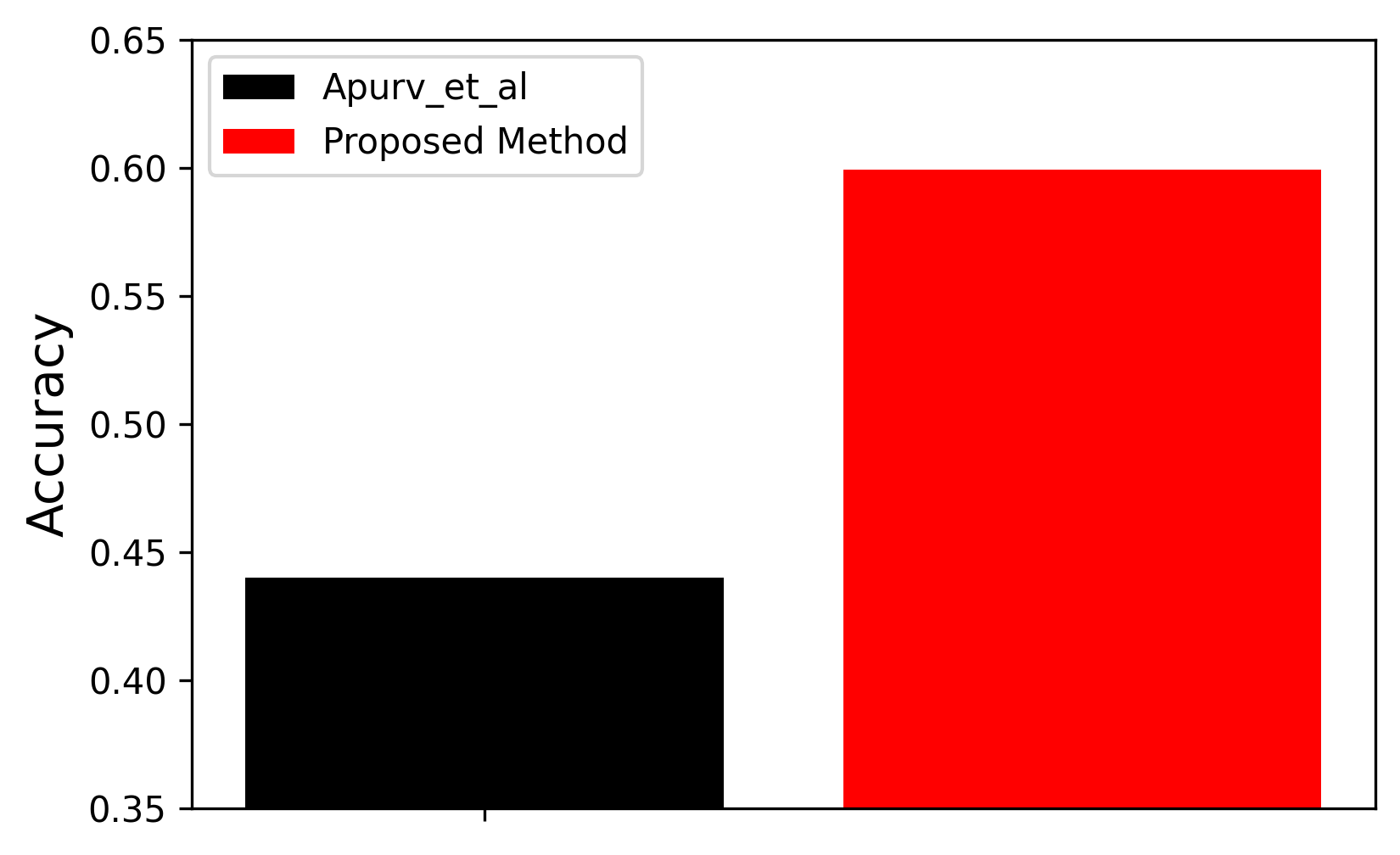}
\end{center}
  \caption{Overall Detection Performance. The proposed e-scooter rider detection network is compared to the current state of the art as described in \cite{apurv2021detection}. Results demonstrate that the proposed detection network achieves an accuracy improvement of 15.93\% over the current state of the art.}
\label{fig:pip2P}
\end{figure}

\begin{figure}[]
\begin{center}
   \includegraphics[width=\linewidth]{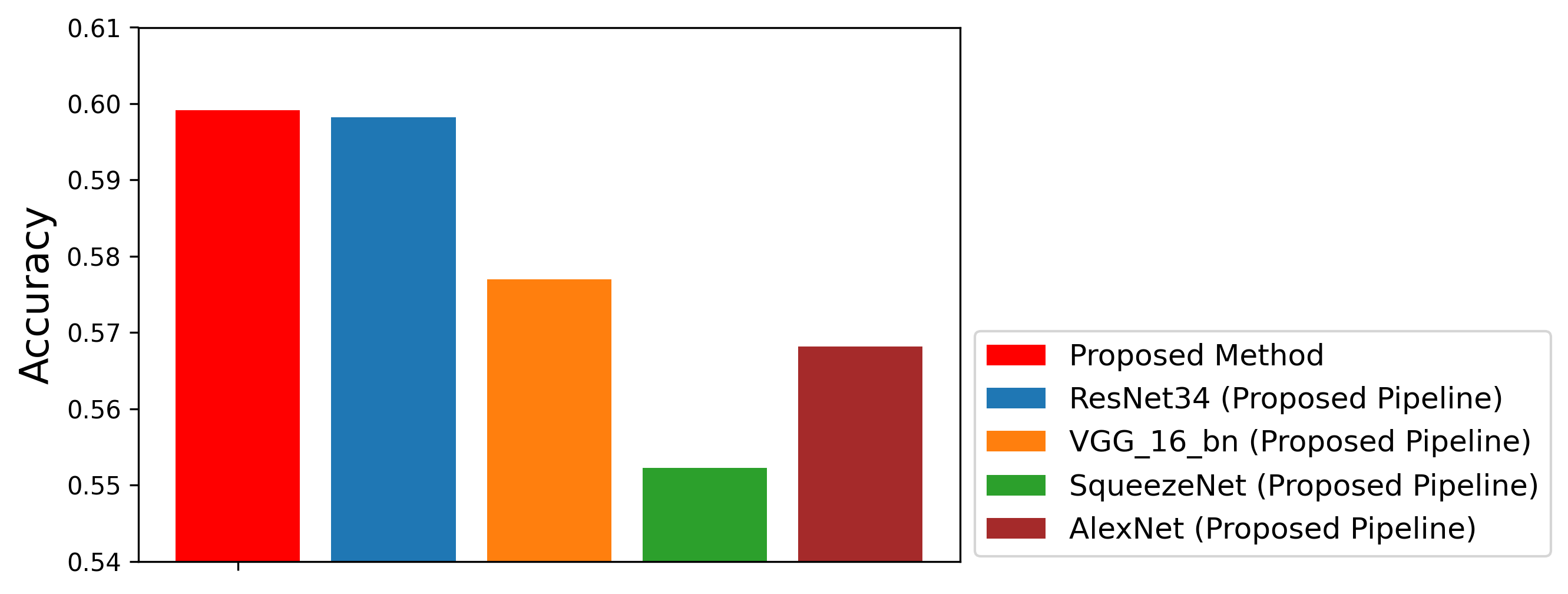}
\end{center}
  \caption{Classifier Comparison using the Proposed Occlusion-Aware Pipeline. The proposed e-scooter rider detection network is compared to four alternative classifier configurations using the proposed pipeline. The ResNet101 classifier specified by the proposed method achieves the highest classification performance with an accuracy of 0.599.}
\label{fig:pip2P_class}
\end{figure}

\begin{figure*}[t]

\subfloat[]{
    \includegraphics[width=\textwidth]{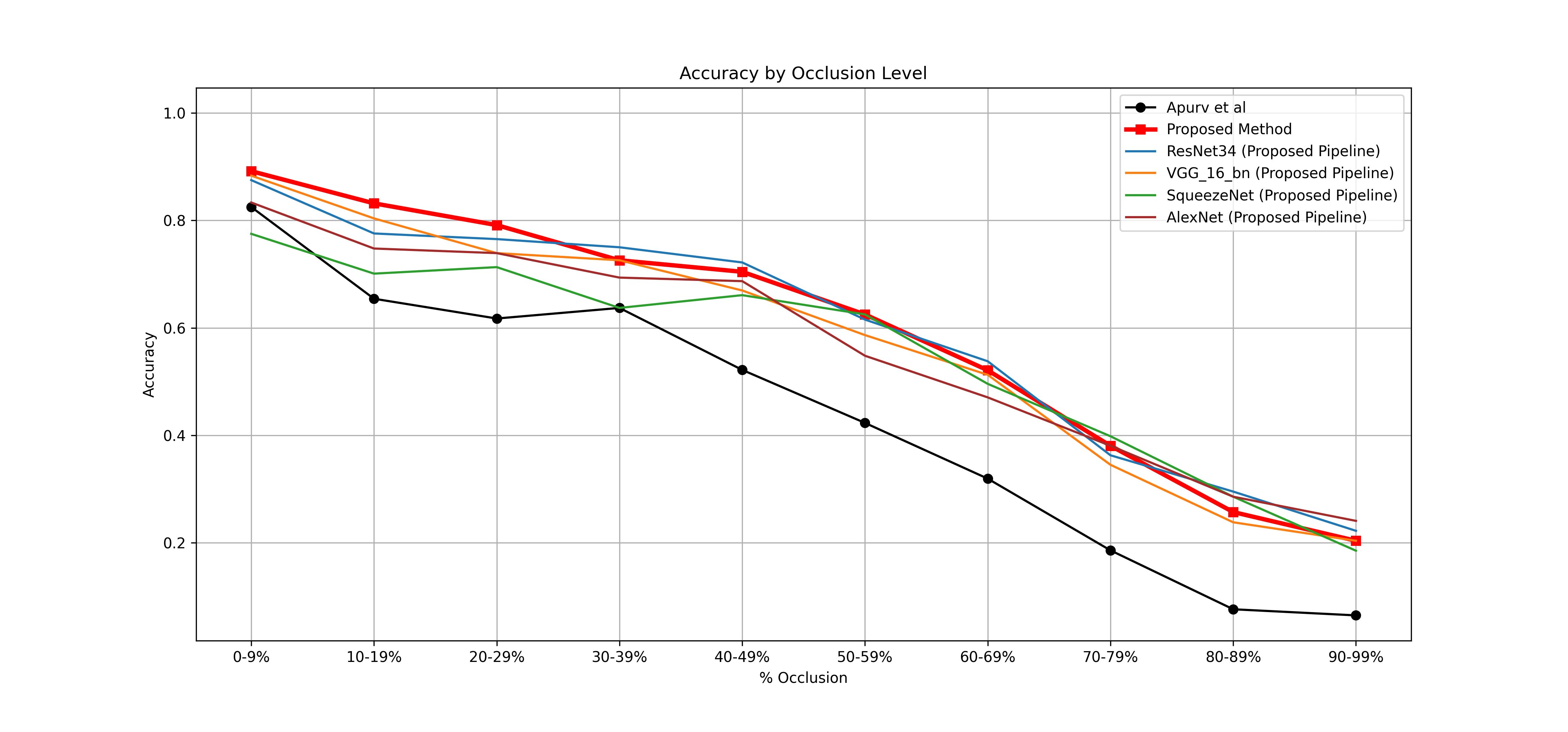}
\label{fig:p2_acc}}

\subfloat[]{
    \includegraphics[width=\textwidth]{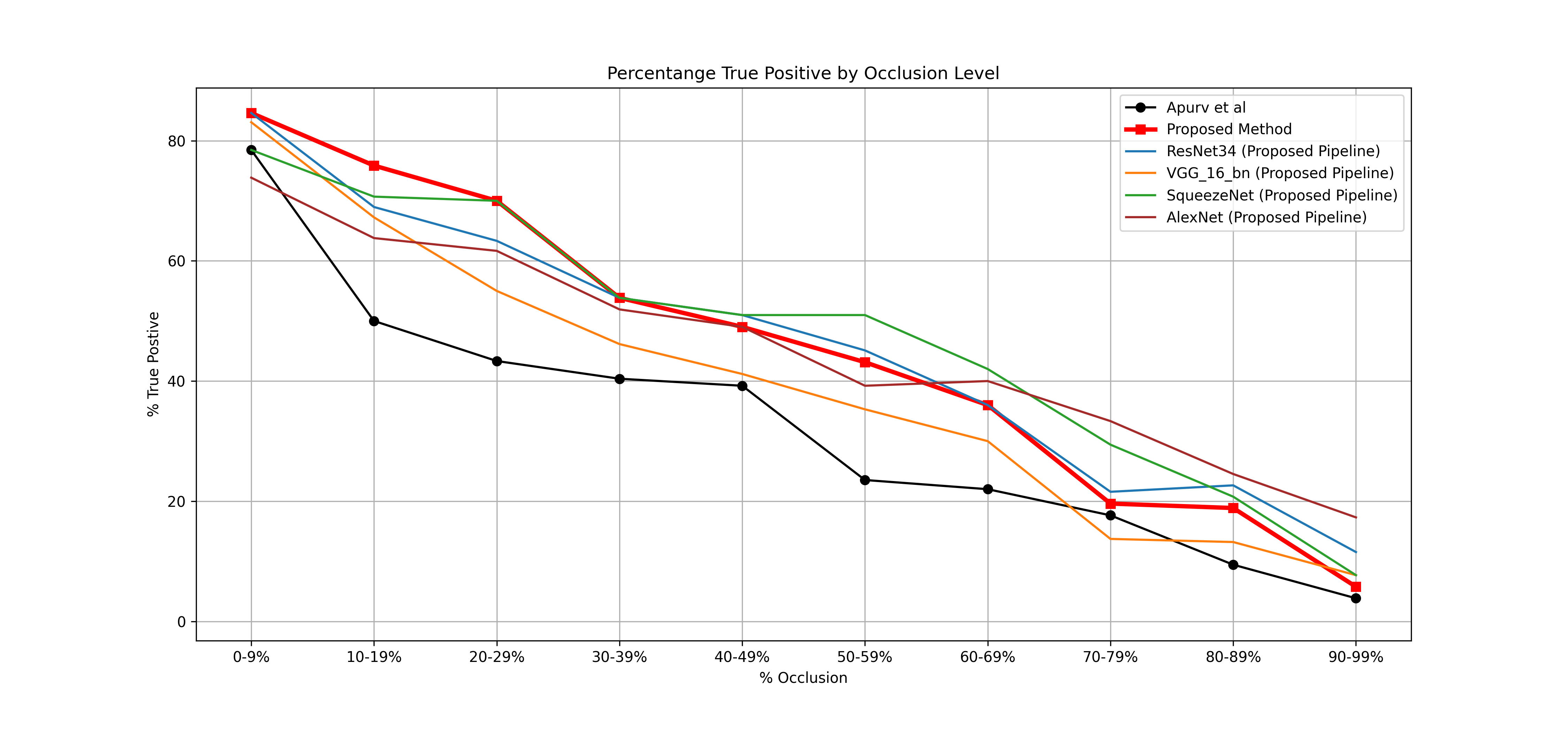}

\label{fig:p2_tp}}



\caption{Detection Performance by Occlusion Level. The detection accuracy by occlusion level, (a), and the percentage of true positives per occlusion level, (b), is shown for the current state of the art, the proposed method and for a number of alternative classifier configurations using the proposed pipeline. The proposed method (red) consistently achieves a higher accuracy and a higher percentage of true positives than the current state of the art (black) \cite{apurv2021detection} across the complete range of occlusion levels.}
\label{occVsdet}
\end{figure*}

\begin{figure*}[t]

    \includegraphics[width=\textwidth]{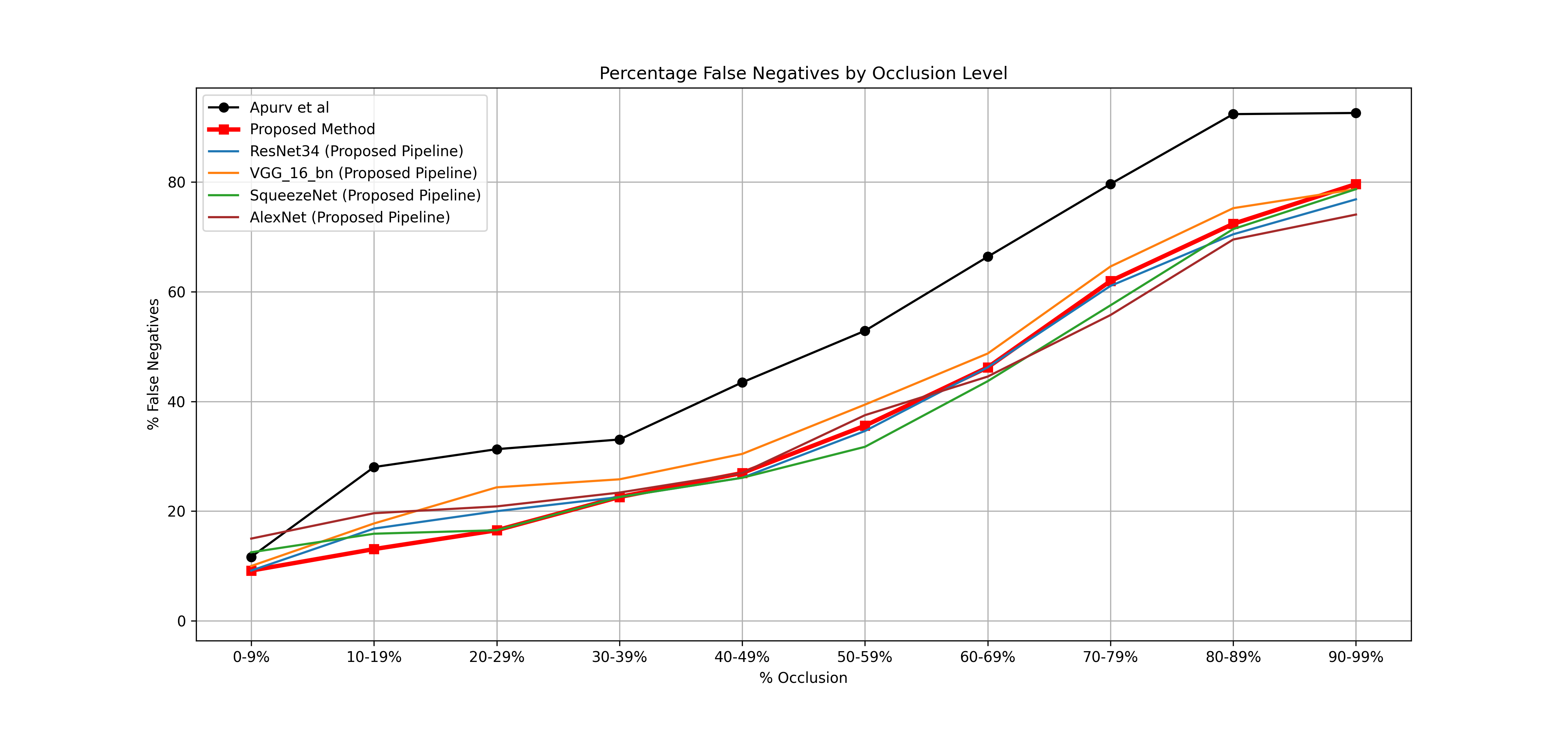}

\caption{False Negatives by Occlusion Level. The percentage of false negatives per occlusion level is shown for the current state of the art, the proposed method and for a number of alternative classifier configurations using the proposed pipeline. The proposed method (red) consistently achieves a lower percentage of false negatives than the current state of the art (black) \cite{apurv2021detection} across the complete range of occlusion levels.}
\label{fig:p2_fn}
\end{figure*}

Detection and classification performance is characterized for e-scooter riders and other vulnerable road users for the complete test dataset and for each level of occlusion from 0-9\% to 90-99\%. The detection method proposed in Section \ref{sec:occ} is compared to the current state of the art \cite{apurv2021detection}, and to four additional classifier configurations based on the proposed pipeline. All classifiers, AlexNet \cite{krizhevsky2014one}, SqueezeNet1.0 \cite{iandola2016squeezenet}, VGG16 with Batch Normalisation (VGG16\_bn) \cite{simonyan2014very}, ResNet34 and ResNet101 \cite{he2016deep}, are trained using the e-scooter rider dataset presented in \cite{apurv2021detection}. 
The overall detection performance of each network can be seen in Figure \ref{fig:pip2P} and Figure \ref{fig:pip2P_class}. Detailed characterization of the detection performance for each level of occlusion is presented in Figure \ref{occVsdet} and Figure \ref{fig:p2_fn}.


\begin{figure}[t]
    
    \includegraphics[width=\linewidth]{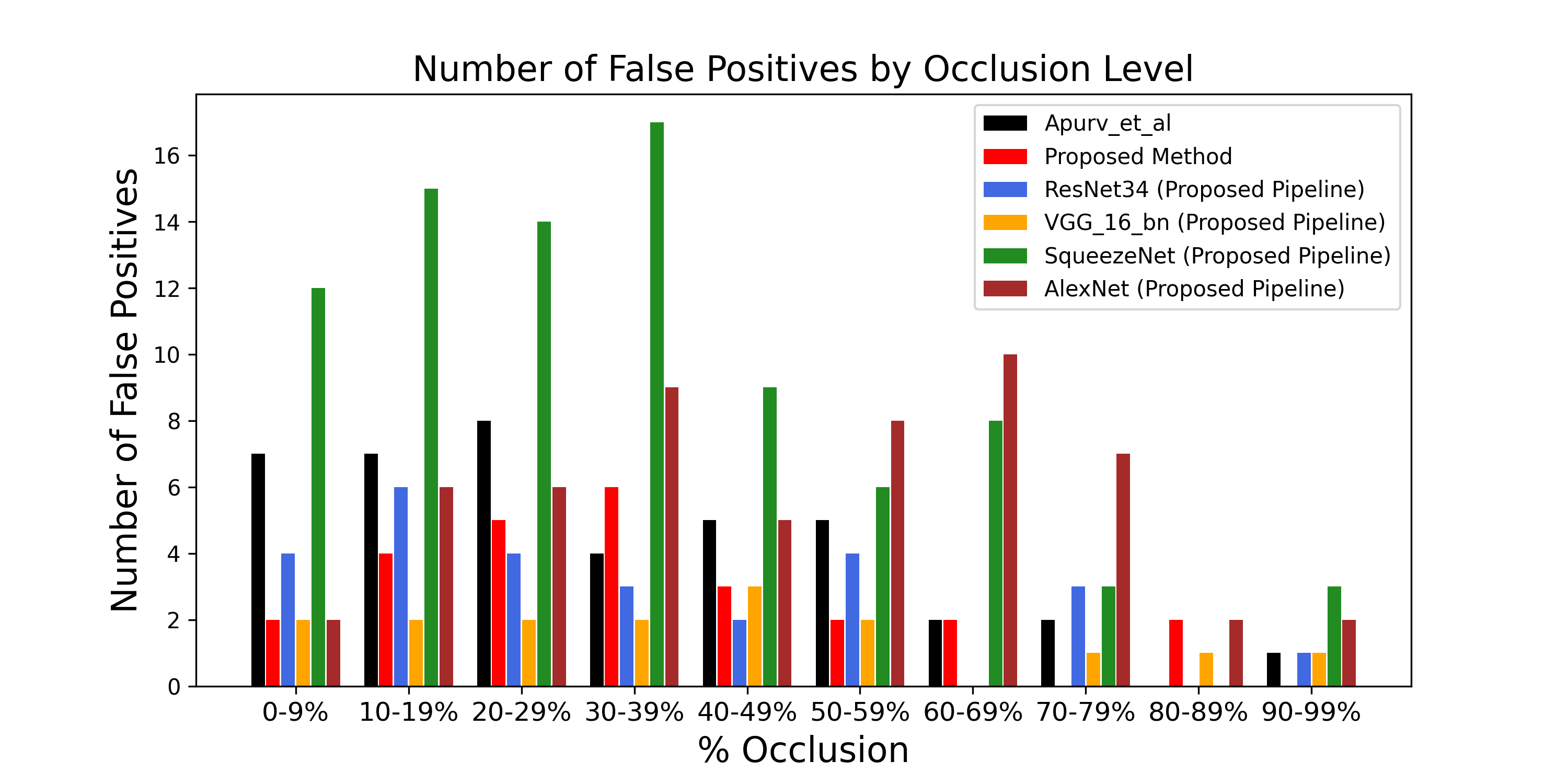}

\caption{Number of False Positives by Occlusion Level. SqueezeNet1.0 \cite{iandola2016squeezenet} detects the highest number of false positives across the range of occlusion levels (87 total false positives), followed by AlexNet \cite{krizhevsky2014one} (57 total false positives).}
\label{fig:p2_fp}
\end{figure}

\section{Results and Analysis}

Figure \ref{fig:yoloP} compares the performance of five popular classification networks based on the methodology outlined by the current state of the art \cite{apurv2021detection}. Results demonstrate that for a mixed occlusion dataset, ResNet101 and ResNet34 \cite{he2016deep} achieve a 2.1\% improvement over the MobileNetV2 classifier \cite{sandler2018mobilenetv2} used by Apurv \textit{et al} \cite{apurv2021detection}, using the same training data, backbone detection network, and classification pipeline.

A novel occlusion-aware method of e-scooter rider detection is described in Section \ref{sec:occ}. Detailed performance characterization for each level of occlusion is carried out for the proposed method, and for a number of alternative classifier configurations, compared to the current state of the art \cite{apurv2021detection}. 
Results demonstrate that the proposed methodology is more proficient at detecting partially occluded e-scooter riders with an overall accuracy of 0.599, a 15.93\% improvement over the current state of the art \cite{apurv2021detection}, Figure \ref{fig:pip2P}. Detailed results of the detection accuracy and the percentage of true positives for each occlusion level are shown in Figure \ref{fig:p2_acc} and Figure \ref{fig:p2_tp} respectively. The number of false negatives by occlusion level is shown in Figure \ref{fig:p2_fn}. Characterization results show that for each level of occlusion, the proposed method provides a superior detection accuracy, a higher percentage of true positives and a lower percentage of false negatives than the current state of the art \cite{apurv2021detection}, Figure \ref{occVsdet} and Figure \ref{fig:p2_fn}. Results also demonstrate that, in general, e-scooter detection accuracy, and the percentage of true positives decline as occlusion level increases, and the percentage of false negatives increase with occlusion level. This reflects the findings of Gilroy \textit{et al} \cite{gilroy2022impact} and presents a significant challenge when detecting and classifying e-scooter riders in dense urban environments where the frequency and severity of partial occlusion is increased. 

Thorough characterization of a detection algorithm at the system development stage can help identify the suitability of specific classification models for particular scenarios and applications. For example, further analysis demonstrates that, although achieving the third most accurate classification performance overall, Figure \ref{fig:pip2P_class}, Vgg16\_bn \cite{simonyan2014very} achieves a below average true positive rate, Figure \ref{fig:p2_tp}, and a slightly above average false negative rate, Figure \ref{fig:p2_fn}, for instances that are more than 10\% occluded. However, VGG16\_bn also maintains a relatively low number of false positive detections across the range of occlusion levels, Figure \ref{fig:p2_fp}. This provides insight into the selectivity of the network and the relatively lower confidence assigned to borderline detection instances.
SqueezeNet1.0 \cite{iandola2016squeezenet} has a higher number of true positive detections for e-scooter riders who are between 40\% and 60\% occluded. AlexNet \cite{krizhevsky2014one} achieves a higher percentage of true positives for instances that are more than 60\% occluded, Figure \ref{fig:p2_tp}. However, both networks incur a significantly higher false positive rate across the range of occlusion levels, Figure \ref{fig:p2_fp}. This is an important distinction as the mis-classification of e-scooter riders as pedestrians or vice versa, can result in dangerous scenarios in autonomous vehicle applications, such as the inappropriate application of emergency braking, potentially resulting in collisions from behind, erratic swerving or the unnecessary triggering of other accident mitigation routines.

\section{Conclusion}
E-scooter usage is predicted to increase in urban environments over the coming decade as cities attempt to reduce congestion, emissions and overcome parking difficulties \cite{heineke2019micromobility}.
The non-detection, or mis-classification of e-scooter users as pedestrians or other road users will have a significant impact on the accident mitigation capabilities and the safe navigation of smart, connected and autonomous vehicles. A large amount of research and benchmarking has been conducted for occluded pedestrian and cyclist detection to date, however, there has been very limited development on e-scooter rider detection in similar scenarios. This research presents an objective test benchmark for the characterization of detection models for partially occluded e-scooter riders. The novel, occlusion-aware e-scooter rider detection method described in this article achieves a 15.93\% improvement in detection accuracy over the current state of the art as presented in \cite{apurv2021detection}. Detailed characterization of the proposed method, and the current state of the art, is provided for the complete range of occlusion levels from 0 to 99\% occluded. 

There is large scope for future work in the field of e-scooter rider detection as this particularly vulnerable class of road user remains largely unrepresented in VRU detection benchmarks. Larger and more diverse training datasets are required to increase detection performance and the ability to generalize to new data. 
High performing single frame e-scooter rider detection algorithms are required to achieve the near real time demands of autonomous vehicle applications, particularly for scenarios such as e-scooter riders emerging from behind parked vehicles or other occlusions. However, frame comparison techniques can be used for less time-sensitive scenarios such as tracking instances that are a further distance from the ego vehicle. Frame comparison based methods will yield more reliable results over multiple frames as e-scooter riders navigate through occlusions of varying severity in heterogeneous traffic. This speed vs. accuracy trade off can be selected at the design stage of an overall detection system, based on the desired application and available sensor configuration. Widespread use of the proposed benchmark will result in more objective, consistent and detailed analysis of detection models for partially occluded e-scooter riders and inform the development of higher performing algorithms across the complete range of occlusion levels, frequently encountered in dense urban environments.

{\small
\bibliographystyle{ieeetr}
\bibliography{egbib}
}

\end{document}